# Customers Behavior Modeling by Semi-Supervised Learning in Customer Relationship Management

[1] Siavash Emtiyaz, [2] MohammadReza Keyvanpour

[1] *Islamic Azad University Qazvin Branch Department of Computer Engineering,*
*Siavash1065@gmail.com*

[2] *Alzahra University Department of Computer Engineering, keyvanpour@alzahra.ac.ir*

## Abstract

*Leveraging the power of increasing amounts of data to analyze customer base for attracting and retaining the most valuable customers is a major problem facing companies in this information age. Data mining technologies extract hidden information and knowledge from large data stored in databases or data warehouses, thereby supporting the corporate decision making process. CRM uses data mining (one of the elements of CRM) techniques to interact with customers. This study investigates the use of a technique, semi-supervised learning, for the management and analysis of customer-related data warehouse and information. The idea of semi-supervised learning is to learn not only from the labeled training data, but to exploit also the structural information in additionally available unlabeled data. The proposed semi-supervised method is a model by means of a feed-forward neural network trained by a back propagation algorithm (multi-layer perceptron) in order to predict the category of an unknown customer (potential customers). In addition, this technique can be used with Rapid Miner tools for both labeled and unlabeled data.*

***Keywords***: *Customer Behavior modeling; Customer Relationship Management; Semi-Supervised Learning; Data Mining.*

## 1. Introduction

For a long time, the focus of modern companies has been shifting from being product oriented to customer-centric organizations. In the industry it is commonly held that maintaining existing customers is more cost-effective than attracting new ones, and that 20% of customers create 80% of the profit [1]. Reichheld and Teal [2] also point out that a 5% increase in customer retention leads to a 25–95% increase in company profit. Therefore, companies are focusing attention on building relationships with their customers in order to improve satisfaction and retention. This implies that companies must learn much about their customers' needs and demands, their tastes and buying propensities, etc., which is the focus of Customer Relationship Management(CRM)[3].For CRM purposes, data mining techniques can potentially be used to extract hidden information from customer databases or data warehouses.CRM includes methodologies and products (e.g. software) that maintain, manage, and optimize the customer's relationship with companies. In short, the goal of CRM is to provide an optimal balance between a company's investment and the customers' satisfaction [2].This study introduces a semi-supervised learning technique to enhance CRM processes and their efficiency. Machine learning algorithms have been shown to be practical methods for real-world recognition problems. They also have proven to be efficient in domains that are highly dynamic with respect to many values and conditions. Various machine learning algorithms have been proposed in the literature [4].This paper explores the possibility of using semi-supervised Learning for analytical CRM. The reality is that the situation for operational CRM application is no different from other real-world problems. In this application, unlabeled data can be collected by automated means from various databases, while labeled data requires human experts or other limited or expensive categorization resources. The fact that unlabelled data is readily available, or inexpensive to collect, can be appealing and one may want to use them. Despite the natural appeal of using unlabeled data, it is not obvious how records without labels can help to develop a system for predicting the labels. This paper introduces a feed forward neural network model, which combines a large set of unlabeled data with a small set of labeled records, to boost the performance of classification. The algorithm is discussed in detail in the following sections.

## 2. Related work

Analytical CRM is gaining momentum these days, and some approaches have been proposed in the literature.Padmanabhan and Tuzhilin [11] categorized CRM components into three groups: adaptive learning, forward looking and optimization. Using these components, they formulated CRM in such a way that state-dependent factors were taken into consideration. Using the proposed framework, they were able to learn the evolution of customer demand, as well as the dynamic effect of its marketing. Xu and Walton [10] have discussed the latest findings of CRM systems and their application. In addition, they introduced an analytical CRM system for customer knowledge acquisition. Eelectronic CRM has also been investigated in [10].In this study[2], extracted hidden information from the logs to automatically tailor ranking functions for a particular user group or collection. Baesens et al. introduced a method with which they were able to predict the customers' future spending patterns. Predicting customer behavior has also been investigated using a transactional database [12].

## 3. Methodology

Machine learning algorithms have been successfully applied in various domains including analytical CRM. Semi-supervised learning (SSL) is halfway between supervised and unsupervised learning. A set of labeled data is used for designing an initial classifier, which is then used for labeling the remaining unlabelled data. Once this is done a classifier is constructed on the basis of both the original and newly labeled data.In addition, it is a well-known fact that when the problem domain is too complex, a large number of training records is required for classification. However, in CRM, similar to other real-world applications, unlabeled data is abundant and readily available, while labeled data requires human experts or other expensive categorization tools. In this paper, we introduce a semi-supervised learning technique that requires only a small size of labeled data for training. Additional labeled records will be collected automatically.

### 3.1 The Data

We examined the proposed algorithm on two datasets. Thiess datasets representing customer relationship management problems have been used in the experiment.First dataset is related to bank dataset. It contained 1000 records and 32 variables. Dataset was divided into a training subset of 700 records, and a test subset of 300 records. Then, the training subset was divided into labeled and unlabelled subsets of 216 and 440 respectively. The training data consist two classes (Good and Bad).
Secondary dataset is related to insurance dataset. It contained 8802 records and 22 variables. Dataset was divided into a training subset of 6000 records, and a test subset of 2802 records. Then, the training subset was divided into labeled and unlabelled subsets of 2250 and 3750 respectively. The training data consist two classes (Yes and No).

### 3.2 Proposed Semi-Supervised Algorithm

In this paper, we introduce a semi-supervised learning method that can be employed in analytical CRM systems. Various semi-supervised learning methods have been proposed. Self-training is a commonly used technique for semi-supervised learning. In self training a classifier is first trained with the small amount of labeled data. The classifier is then used to classify the unlabeled data. Typically the most confident unlabeled points, together with their predicted labels, are added to the training set. The classifier is re-trained and the procedure repeated. Note the classifier uses its own predictions to teach itself. The procedure is also called self-teaching or bootstrapping (not to be confused with the statistical procedure with the same name) [7].
Another popular approach is co-training. Co-training assumes that:
   (i)   features can be split into two sets
   (ii)  Each sub-feature set is sufficient to train a good classifier
   (iii) The two sets are conditionally independent given the class.

Initially two separate classifiers are trained with the labeled data, on the two sub-feature sets respectively. Each classifier then classifies the unlabeled data, and 'teaches' the other classifier with the few unlabeled examples (and the predicted labels) they feel most confident. Each classifier is retrained with the additional training examples given by the other classifier, and the process repeats. In co-training, unlabeled data helps by reducing the version space size. In other words, the two classifiers (or hypotheses) must agree on the much larger unlabeled data as well as the labeled data. In this paper, we use of self-training algorithm for customer behavior modeling [7]. (Figure 1)

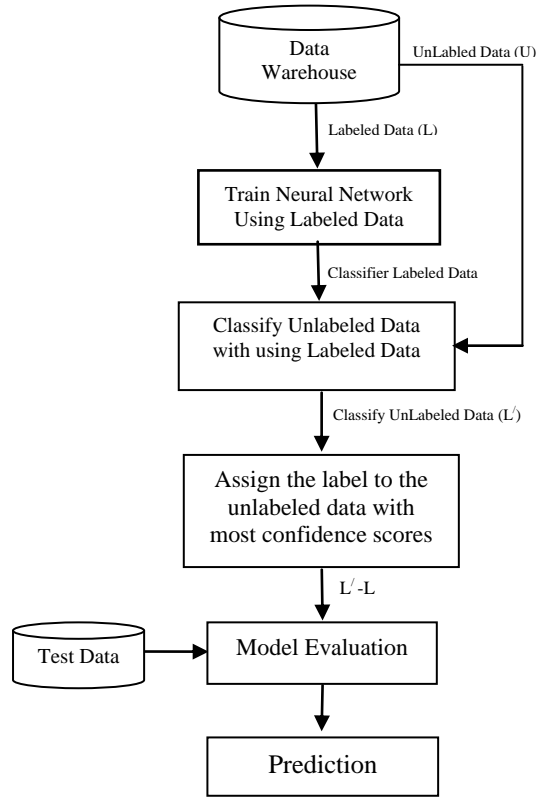

**Figure 1.** An overview of the Algorithm

The proposed method is multi-layer perceptron neural network algorithm in which a back propagation (BP) algorithm [10] is used to classifier. Multi-Layer Perceptron networks consists of multiple layers of computational units, usually interconnected in a feed-forward way. Each neuron in one layer has directed connections to the neurons of the subsequent layer. In many applications the units of these networks apply a sigmoid function as an activation function. Each layer consists of units which receive their input from units from a layer directly below and send their output to units in a layer directly above the unit. There are no connections within a layer [9,14].

Multi-layer networks use a variety of learning techniques, the most popular being back-propagation. Here, the output values are compared with the correct answer to compute the value of some predefined error-function. By various techniques, the error is then fed back through the network. Using this information, the algorithm adjusts the weights of each connection in order to reduce the value of the error function by some small amount. After repeating this process for a sufficiently large number of training cycles, the network will usually converge to some state where the error of the calculations is small. In this case, one would say that the network has learned a certain target function. To adjust weights properly, one applies a general method for non-linear optimization that is called gradient descent[14].Consider a network with a single real input x and network function F. The derivative F'(x) is computed in two phases:

- Feed Forward: the input x is fed into the network. The primitive functions at the nodes and their derivatives are evaluated at each node. The derivatives are stored.

- Back propagation: the constant 1 is fed into the output unit and the network is run backwards. Incoming information to a node is added and the result is transmitted to the left of the unit. The result collected at the input unit is the derivative of the network function with respect to x [9,14].

$$E(\vec{w}) \equiv \frac{1}{2} \sum_{d \in D} \sum_{k \in outputs} (t_{kd} - O_{kd})^2 \qquad (1)$$

In This Paper, We define the structure of the neural network with the parameter list hidden layers. Each list entry describes a new hidden layer. The key of each entry must correspond to the layer name [10]. The value of each entry must be a number defining the size of the hidden layer. A size value of -1 indicates that the layer size should be calculated from the number of attributes of the input example set. In this case, the layer size will be set to (Number of attributes + Number of classes) / 4.If do not specify any hidden layers, a default hidden layer with sigmoid type and size (number of attributes + number of classes) / 4 will be created and added to the net. If only a single layer without nodes is specified, the input nodes are directly connected to the output nodes and no hidden layer will be used. The used activation function is the usual sigmoid function. Therefore, the values ranges of the attributes should be scaled to -1 and +1.The type of the output node is sigmoid if the learning data describes a classification task and linear for numerical regression tasks[9].

## 4. Experiment and Result

The proposed semi-supervised algorithm was carried out using Rapid Miner 5.1 (http://www.rapid-i.com). Rapid Miner is unquestionable the world-leading open-source system for data mining. It is available as a stand-alone application for data analysis and as a data mining engine for the integration into own products. It provides data mining and machine learning procedures including: data loading and transformation (ETL), data preprocessing and visualization, modeling, evaluation, and deployment. Design environment contain three primary operators:

- Neural network: This operator learns a model by means of a feed-forward neural network trained by a back propagation algorithm.
- Apply Model: This operator applies a model to unlabeled data. Models usually contain information about the data they have been trained on. This information can be used for predicting the value of a possibly unknown label. (Table 1)
- Performance: This operator delivers list of performance values automatically determined in order to fit the learning task type.

| *Raw* No | *Confidence*(no response) | *Confidence*( response) | *Assign* Labeled Data |
|---|---|---|---|
| 1 | 0.185 | 0.815 | response |
| 2 | 0.015 | 0.985 | response |
| 3 | 0.937 | 0.063 | no response |
| 4 | 1.000 | 0.000 | no response |
| 5 | 1.000 | 0.000 | no response |
| 6 | 0.144 | 0.856 | response |
| 7 | 0.951 | 0.049 | no response |
| 8 | 1.000 | 0.000 | no response |
| 9 | 0.858 | 0.142 | no response |
| 10 | 0.000 | 1.000 | response |
| 11 | 0.937 | 0.063 | no response |
| 12 | 0.000 | 1.000 | response |

**Table 1.** An overview labeled data to unlabeled data with most confidence

In addition, Design environment contains two secondary operators:

- Nominal to Numerical: This operator maps all non numeric attributes to real valued attributes. Nothing is done for numeric attributes, binary attributes are mapped to 0 and 1. In this case, we maps name attribute to real attribute.
- Set Role: This operator can be used to change the role of an attribute of the input *Example Set*. In this case, we change Label attribute for prediction.

In the test presented here, we split data as random 30% for test and 70% for training. Furthermore, we use confusion matrix for compare the result. A confusion matrix contains information about actual and predicted classifications done by a classification system. Performance of such systems is commonly evaluated using the data in the matrix. The table 2 shows the confusion matrix for a two class classifier.

The entries in the confusion matrix have the following meaning in the context of our study:
- a is the number of correct predictions that an instance is negative,
- b is the number of incorrect predictions that an instance is positive,
- c is the number of incorrect of predictions that an instance negative, and
- d is the number of correct predictions that an instance is positive

**Table 2.** confusion matrix for two class classifier[13]

|  |  | Predicted | |
|---|---|---|---|
|  |  | **Negative** | **Positive** |
| *Actual* | **Negative** | a | b |
|  | **Positive** | c | d |

- The accuracy (AC) is the proportion of the total number of predictions that were correct. It is determined using the equation: $AC = \frac{a+b}{a+b+c+d}$

- The recall or true positive rate (TP) is the proportion of positive cases that were correctly identified, as calculated using the equation: $TP = \frac{d}{c+d}$

- The false positive rate (FP) is the proportion of negatives cases that were incorrectly classified as positive, as calculated using the equation: $FP = \frac{b}{a+b}$

- The true negative rate (TN) is defined as the proportion of negatives cases that were classified correctly, as calculated using the equation: $TN = \frac{a}{a+b}$

- The false negative rate (FN) is the proportion of positives cases that were incorrectly classified as negative, as calculated using the equation: $FN = \frac{c}{c+d}$

We trained one-hidden layer of neural net operator in Rapid Miner with sigmoid of hidden units according following formula:((number of attributes + number of classes) / 2) + 1).Furthermore, using the error-back propagation algorithm, training cycle 500, training rate 0.3, and error epsilon 1.0E-5.
Rapid Miner is open source system for data mining and we modified this operator codes and we could get a better result than the previous. We modified trained with one-hidden layer of neural net operator in Rapid miner with sigmoid of hidden units according following formula: (Number of attributes + Number of classes) / 4.
In figure 2, we observe a plot comparison to show the percent error in the performance of neural network algorithm for both modified neural network and neural network. Therefore, we have used above the formula for this research. Blue line is modified neural network
which is defined by the following formula:
    (Number of attributes + Number of classes) / X    x=1,2,3,4,5,6
Also, the red line is the former neural network which is defined by the following formula:
    ((Number of attributes + Number of classes) / X)+1    x=1,2,3,4,5,6

As can be seen in the figure 2, in parts three and four our algorithm has the lowest percentage of error. Results are presented in table 3. In table 3 we show test result on the bank data set. These results have been 500 training cycle. In figure 3 we show test performance(%) for proposed semi-supervised algorithm on the bank data set. Also, in figure 3 we observe a significant increase in the performance of proposed semi-supervised. Also, results are presented in Table 4,In Table 4 we show test result on the insurance data set.

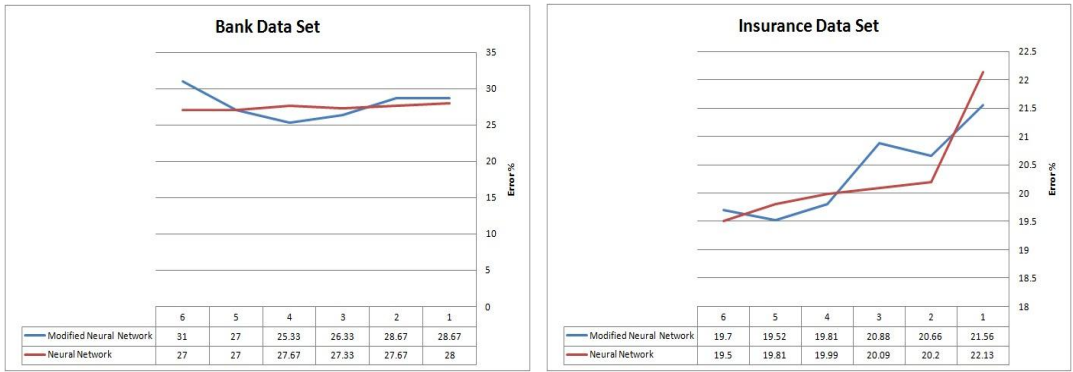

**Figure 2.** a plot comparison to show the percent error in bank data set and insurance data set

**Table 3.** Test Performance (%) for proposed semi-supervised algorithm on the bank data set

| *Operator* | *Accuracy*(%) | *True* **Good** (%) | *True* **Bad** (%) | *False* **Good** (%) | *False* **Bad** (%) |
|---|---|---|---|---|---|
| Neural Net | 72.33 | 79.25 | 55.68 | 44.32 | 20.75 |
| Proposed algorithm | 74.67 | 80.18 | 60.24 | 39.76 | 19.82 |

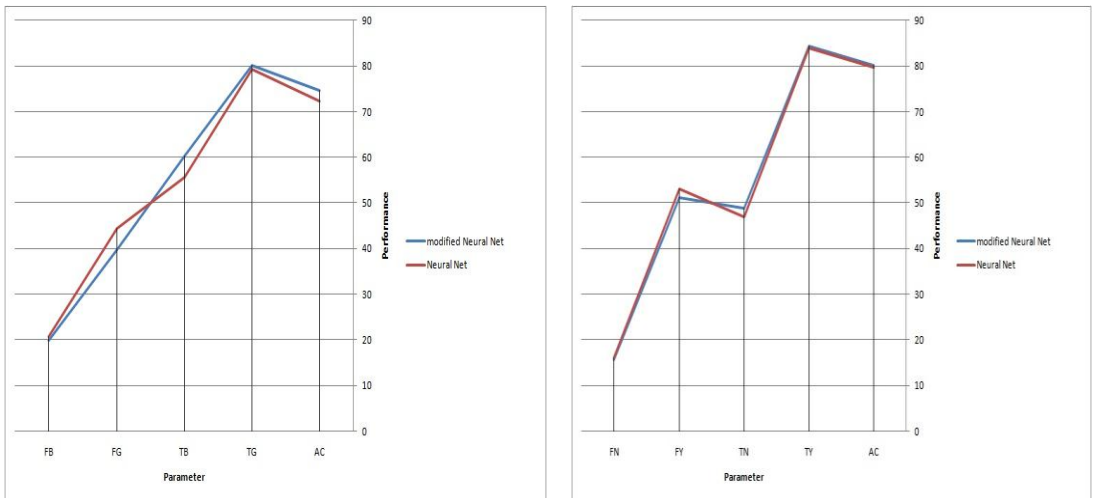

**Figure 3.** Test performance on the bank data set(Left) and Test performance on the insurance data set(Right)

**Table 4.** Test Performance(%) for proposed semi-supervised algorithm on the Insurance data set

| *Operator* | *Accuracy*(%) | *True* **Yes** (%) | *True* **No** (%) | *False* **Yes** (%) | *False* **No** (%) |
|---|---|---|---|---|---|
| Neural Net | 79.80 | 83.98 | 47.00 | 53.00 | 16.02 |
| Proposed algorithm | 80.19 | 84.44 | 48.80 | 51.20 | 15.56 |

We trained the neural network classifier with the training data and tested on records. The lowest error we achieved 5.33%. The proposed algorithm illustrated similar generalization capability to the classifier trained with the original training records, with the difference of approximately 4%.

Now, we want compare our proposed semi-supervised algorithm with other algorithms in rapid miner. Result are presented in table 5 and table 6 on the bank and insurance data sets respectively.

## 5. Conclusion & Future Work

The most important results obtained from the above techniques are reorganization of data and information for easier access, yielding more effective Efficiency and better data classification in order to obtain best results from them. Data mining and CRM integration has advantages and lots
of benefits for companies that need to discover profitability of some customers than other customers. Data mining can identify important customers in data warehouse.
Our approach has follow advantages: firstly, semi-supervised machine learning techniques are used to automatically construct customer behavior modeling to improve accuracy. Secondary, a neural network can be used for data visualization purposes. Data visualization is an important step in data analysis and can provide valuable insights into all steps of classification.
Future work will also examine the use other classification methods, such as fuzzy neural network, to classifier labeled data of varying confidences. We are also interested in benchmarking our method on other operational CRM databases.

**Table 5.** Test Performance (%) for proposed semi-supervised algorithm and other algorithms on the bank data set

| *Operator* | *Accuracy* (%) | *True* **Good** (%) | *True* **Bad** (%) | *False* **Good** (%) | *False* **Bad** (%) |
|---|---|---|---|---|---|
| Proposed algorithm | 74.67 | 80.18 | 60.24 | 39.76 | 19.82 |
| Neural Net | 72.33 | 79.25 | 55.68 | 44.32 | 20.75 |
| SVM | 73.67 | 79.91 | 58.14 | 41.86 | 20.09 |
| KNN | 59.67 | 70.48 | 34.44 | 65.56 | 29.52 |
| Naive Bayes | 55.67 | 79.84 | 38.64 | 61.36 | 20.16 |

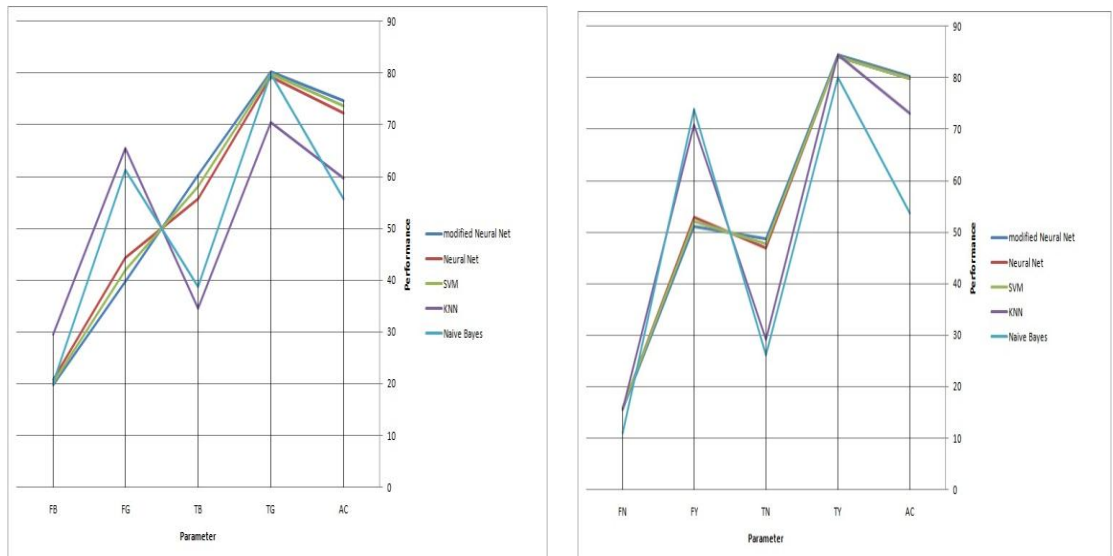

**Figure 4.** Test performance on the Bank data set(Left) And Insurance dataset(Right) with other algorithms

**Table 6.** Test Performance (%) for proposed semi-supervised algorithm and other algorithms on the insurance data set

| Operator | Accuracy(%) | True Yes (%) | True No (%) | False Yes (%) | False No (%) |
|---|---|---|---|---|---|
| Proposed algorithm | 80.19 | 84.44 | 48.80 | 51.20 | 15.56 |
| Neural Net | 79.80 | 83.98 | 47.00 | 53.00 | 16.02 |
| SVM | 79.87 | 84.00 | 47.70 | 52.14 | 15.98 |
| KNN | 73.09 | 84.43 | 29.09 | 70.91 | 15.57 |
| Naive Bayes | 53.64 | 80.05 | 26.17 | 73.83 | 10.95 |